\RequirePackage{fix-cm}
\documentclass[twocolumn]{svjour3}          
\smartqed  
\usepackage{graphicx}
\usepackage{amsfonts}
\usepackage{url}
\usepackage{amsmath}
\usepackage{cite}
\usepackage{amsmath}
\usepackage{graphicx}
\usepackage{tabularx}
\usepackage{float}
\usepackage{booktabs}
\usepackage{multirow}
\usepackage{lipsum}
\journalname{CGI2018} 
\begin{document}

\title{GRC-Net: Gram Residual Co-attention Net for epilepsy prediction}

\author{Bihao You \and Jiping Cui}

\institute{Bihao You \at
              School of Advanced Technology, Xi'an Jiaotong-Liverpool University, Suzhou, 215123, China \\
              \email{Bihao.You22@student.xjtlu.edu.cn}
           \and
           Jiping Cui \at
              School of Advanced Technology, Xi’an Jiaotong-Liverpool University, Suzhou, 215123, China;\\
              and also with the School of Electrical Engineering, Electronics and Computer Science, University of Liverpool, Liverpool, L69 3BX, United Kingdom \\
              \email{Jiping.Cui24@student.xjtlu.edu.cn}
}

\date{ }
\maketitle

\begin{abstract}
Prediction of epilepsy based on electroencephalogram (EEG) signals is a rapidly evolving field. Previous studies have traditionally applied 1D processing to the entire EEG signal. However, we have adopted the Gram Matrix method to transform the signals into a 3D representation, enabling modeling of signal relationships across dimensions while preserving the temporal dependencies of the one-dimensional signals. Additionally, we observed an imbalance between local and global signals within the EEG data. Therefore, we introduced multi-level feature extraction, utilizing coattention for capturing global signal characteristics and an inception structure for processing local signals, achieving multi-granular feature extraction. Our experiments on the BONN dataset demonstrate that for the most challenging five-class classification task, GRC-Net achieved an accuracy of 93.66\%, outperforming existing methods.
\end{abstract}

\section{Introduction}
\label{sec:1}
EEG signals are indispensable for analyzing brain activity, understanding brain function, and exploring cognitive mechanisms. Therefore, effective processing of EEG signals is crucial in various domains such as speech recognition \cite{chen2012towards}, emotion detection \cite{petrantonakis2009emotion}, and diagnosing diseases related to brain function \cite{cai2018reconstruction} \cite{oh2019deep} \cite{frantzidis2014functional}. While significant progress has been made in using one-dimensional EEG signal processing for predicting epileptic seizures, there is a growing interest in transforming one-dimensional EEG signal sequences into two-dimensional images to extract deep features from multiple channels for prediction.

\begin{figure}[H]
    \centering
    \includegraphics[width=1\columnwidth]{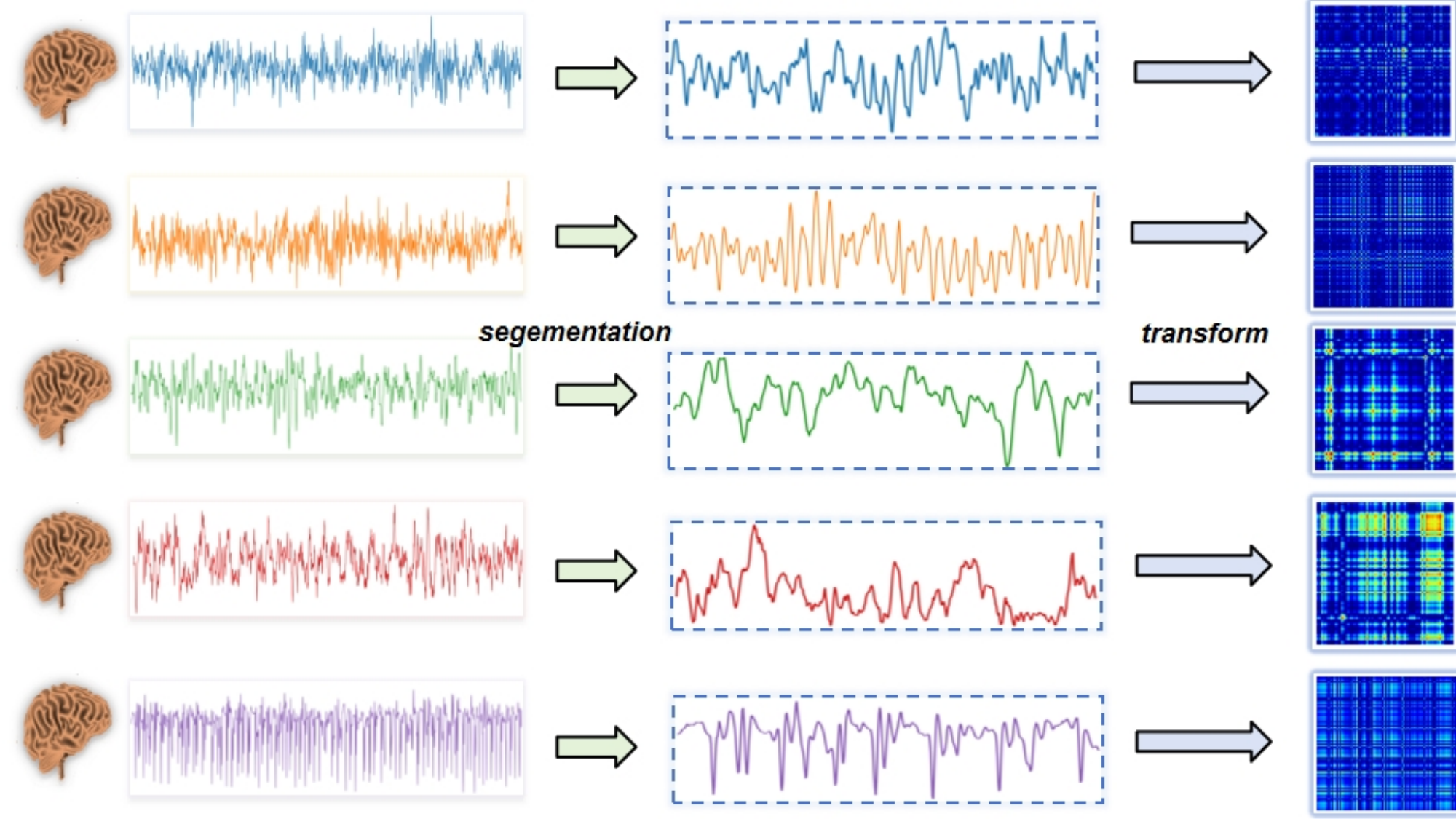}
    \caption{In the first column of images, the original one-dimensional EEG signals from the BoNN dataset are displayed. In the second column, the signals underwent resampling using sliding windows with specified strides. In the third column, the resampled signals were transformed into feature maps using the Gram Matrix method for standardization, while preserving temporal dependencies.}
    \label{fig:Signal}
\end{figure}

With the emergence of EEG datasets for epileptic analysis, such as BONN, several methods have been proposed for seizure prediction. Although these works identified some issues in using EEG signal processing for seizure prediction, their models addressed them in a rudimentary manner.

Previous studies on EEG signal prediction have mostly relied on one-dimensional time series or simply utilized time series to plot corresponding curves. However, we argue that due to the unidirectional correlation in the horizontal and vertical directions of one-dimensional signals, they may not fully capture the underlying relationships in the data. Moreover, the linear operations on one-dimensional sequences fail to discriminate between the meaningful information in the signal and Gaussian noise. Therefore, the utilization of one-dimensional signals alone for plotting corresponding images may lead to a loss of temporal dependencies in the data, resulting in information loss.

Furthermore,traditional CNN-based models lack the ability to model long-range dependencies and perceptions due to their focus on modeling local information. While some researchers have recognized this issue and attempted to incorporate attention mechanisms in models for long-range global modeling, the original Self-Attention structure in Transformers calculates attention matrices based on interactions between queries and keys, thereby overlooking the relationships between adjacent keys. This limitation still hampers the effectiveness of global modeling.

To tackle the primary concern, we propose a technique that converts the original EEG signal based on Gramian Angular Feild(GAF). The GAF method is employed to map one-dimensional EEG signal data onto a polar coordinate system, where as the sequence progresses over time, the corresponding values are distorted between different angular points on the circle. This transformation generates a bidirectional sequence that retains all information without any loss and preserves temporal dependencies. Specifically,as shown in Fig\ref{fig:Signal},we segment the raw data using a sliding window with a specific step size, followed by normalization and signal transformation. Furthermore, to fully leverage the contextual information of the key, we utilize Cotattention. Initially, a 3x3 convolution is applied to the key to model static contextual information. Subsequently, the concatenated output of the query and the modeled key with context information undergoes self-attention using two consecutive 1x1 convolutions to generate dynamic context. The static and dynamic contextual information is ultimately fused to produce the output.
Overall,our contributions can be summarized as follows.
\begin{enumerate}
\item 
We utilized the Gram Matrix to perform a two-dimensional transformation on the one-dimensional signal, thereby avoiding the loss of temporal dependency information during the conversion process.
\item
We employ the ResNet with the Cotattention mechanism, we enhanced the internal static contextual information within the attention mechanism, enabling the modeling of both local and global information.

\item We conduct extensive experiments on BONN dataset, and our GRC-Net achieves state-of-the-art performance.

\end{enumerate}
\section{Related Work}
\label{sec:2}
With the success of Deep Neural Networks in computer vision tasks, improvements have been made in epilepsy prediction relying on the analysis of Electroencephalogram (EEG) signals through deep learning \cite{Tuncer2021EpilepsyAR,Gao2021GenerativeAN,Sivasaravanababu2021AnEE,Mishra2022ADB,Thuwajit2022EEGWaveNetMC,Du2024ElectroencephalographicSD}. EEG signals, unlike other signals, are a type of non-stationary stochastic signal that lacks ergodicity and are characterized by strong background noise. Some methods employ Residual Networks to extract signal features and then analyze the temporal relationships using Long Short-Term Memory networks \cite{Singh2023AutomaticPO,Lee2024ARH}. To address the limitation of 2D CNNs in extracting high-quality discriminative features between channels \cite{jia2022efficient,wusthoff2009limitations}, a method has been proposed to reconstruct time-series EEG signals into 3D feature maps and utilize a 3D-2D Hybrid Convolutional Neural Network (HyCNN) model \cite{Qi2023PredictingES} to extract correlated features among multiple EEG channels. Additionally, Panda et al. \cite{Panda2023HybridWO} presented a hybrid Water Cycle Algorithm (WCA) - Particle Swarm Optimization (PSO) optimized Ensemble Extreme Learning Machine (EELM) for classifying epileptic seizures to improve classification performance. Swetha et al. \cite{Swetha2023OppositionalCS} employed an Artificial Neural Network (ANN) based on the Oppositional Crow Search Algorithm (OCSA) for classifying epileptic seizure disorders. Chou et al. \cite{Chou2023ConvolutionalNN} developed a video EEG-based rapid epileptic seizure detection system using Convolutional Neural Networks. Qin et al. \cite{QIN2023104644} applied an EEG signal recognition framework based on an improved Variational Mode Decomposition (VMD) and Deep Forest.

Recently, there has been a proliferation in approaches based on Short-Time Fourier Transform for extracting spatial and temporal information from EEG signals. For instance, Zhou et al. \cite{ZHOU2024108086} proposed a method for automatic seizure detection called Residual-based Inception with Hybrid-Attention Network (RIHANet). This method initially utilizes Empirical Mode Decomposition and Short-Time Fourier Transform (EMD-STFT) for data processing to enhance the quality of time-frequency representation of EEG signals. Lu et al. introduced the CBAM-3D CNN-LSTM model \cite{Lu2023AnES}, which employs Short-Time Fourier Transform to extract time-frequency planes from time-domain signals for epilepsy classification. Additionally, Kantipudi et al. \cite{Kantipudi2024AnIG} proposed an improved GBSO-TAENN model, which employs Finite Linear Haar Wavelet Filtering (FLHF) technique for filtering input signals, and utilizes Grasshopper Biologically Swarm Optimization (GBSO) technique to select optimal features by computing the best fitness value, and employs Time-Activated Expandable Neural Network (TAENN) mechanism for classifying EEG signals.

Based on the above survey, most previous studies have been based on extracting features from the one-dimensional pulse wave signal and passing these features to a classification model. However, the extracted features did not preserve the temporal dependency of the pulse wave signal, resulting in lower accuracy. Other studies [17] addressed this issue by directly passing the pulse wave signal to an LSTM network. However, LSTM networks have a long training time. While some have converted one-dimensional signals into two-dimensional signals and enhanced performance using networks combined with attention mechanisms, they did not avoid the loss of temporal dependency and the basic attention architecture had limited capabilities for global modeling. In contrast, we mapped the EEG signal to a two-dimensional image using the GM, preserving the temporal dependency of the EEG signal. Additionally, leveraging cotattention in the GRC-Net enhanced the ability for global modeling, fundamentally addressing these challenges.
\section{Methodology}
\subsection{Overview}

\begin{figure*}[!t]
    \centering
    \includegraphics[width=1\textwidth]{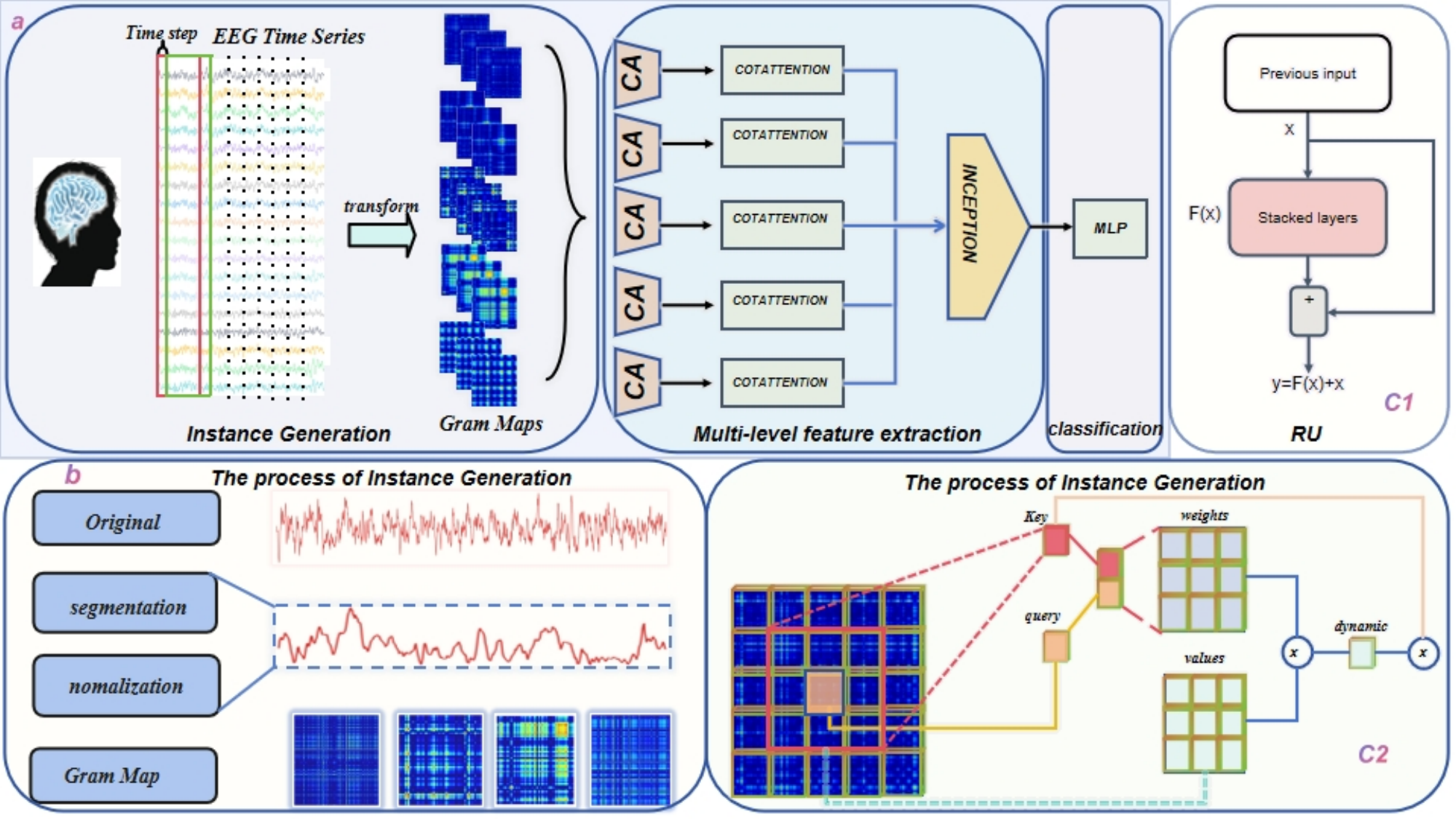}
    \caption{Overview of the proposed GRC-Net framework. (a) The pipeline of GRC-Net includes Instance Generation, Multi-Level Feature Extraction, Fusion, and Classification. (b) A novel approach for transforming one-dimensional signals into two-dimensional Gram Matrix representation. (c) A simplified illustration of Multi-level feature extraction,CA represents channel augument,cotattention is the key component of global feature extraction,RU denotes for residual unit is for local feature extraction.}
    \label{fig:Overview}
\end{figure*}

The GRC-Net proposed by us mainly consists of four steps: instance generation, feature extraction, feature aggregation, and classification. In the instance generation step, through the transformation by the GAF, the one-dimensional signal is converted into a two-dimensional image while preserving temporal dependency features. In the feature extraction stage, the GRC-Net, based on the ResNet architecture, leverages spatial locality and translational equivariance to learn short-range temporal relationships and models long-range temporal relationships through cotattention mechanism. Feature aggregation utilizes the Inception architecture to avoid gradient explosion and classifies the signals through MLP. An overview of the GRC-Net framework used is depicted in Figure \ref{fig:Overview}.


\subsection{Instance Generation}
To generate multiple instances from a single record, we adopted a sliding window approach. A window size of 512 and a stride of 128 (with 25\% overlap) were selected. Each signal of length 4097 in the training set was divided into 57 sub-signals, with each sub-signal considered as an independent signal instance (SI). Consequently, a total of 5700 instances were created for each class.We partitioned the available signal into non-overlapping training and testing sets, each accounting for 90\% and 10\% of the total signal, respectively.

Given a one-dimensional signal instance$SI = \{s_1, \ldots, s_{57}\}$, where the time series consists of 57 timestamps \( t \) and corresponding actual observed values \( s \), we scale the time series to the interval \([-1, 1]\) using the Min-Max scaler, as shown in Equation (1):
\begin{equation}
{ s_i' = \frac{(s_i - \text{max}(SI)) + (s_i - \text{min}(SI))}{\text{max}(SI) - \text{min}(SI)}}    
\end{equation}

\begin{figure}[H]
    \centering
    \includegraphics[width=1\columnwidth]{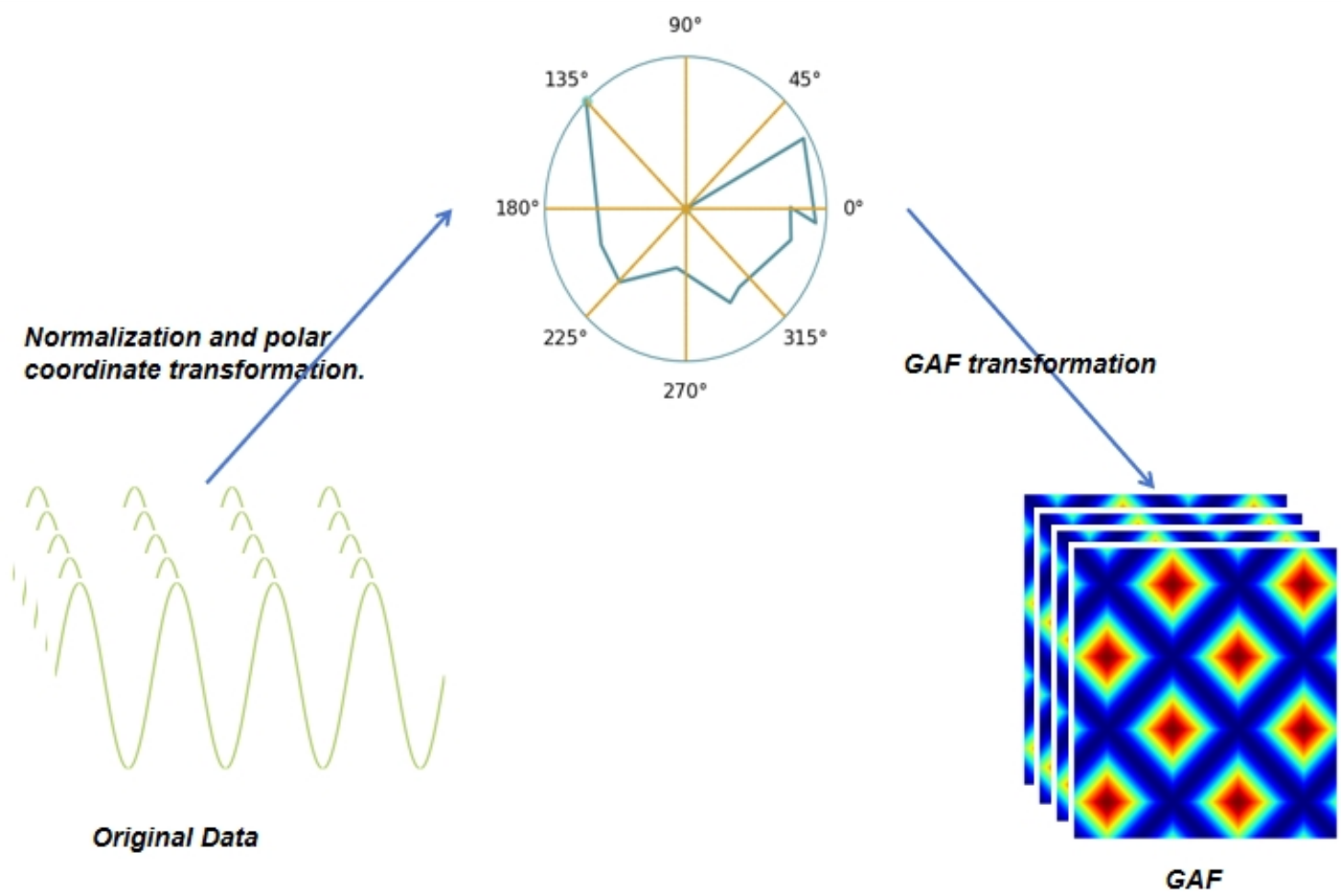}
    \caption{The process of generating GAF involves mapping one-dimensional EEG signal data onto polar coordinates, where the sequence evolves over time causing the values to distort between different focal points on the circle, akin to ripples in water. This polar coordinate-based representation offers a novel approach to understanding one-dimensional signals. Subsequently, by employing angular perspective, GAF is generated by altering spatial dimensions, utilizing angular relationships to uncover hidden characteristics of EEG signals, and the resulting coefficient output makes it easy to distinguish valuable information from Gaussian noise.}
    \label{fig:GAF}
\end{figure}

Subsequently, we map the scaled sequence values $s_i$ to angular values $\theta$ and the times are mapped to radius $r$, allowing the scaled time series to be represented in polar coordinates, as shown in Equation (2,3):

\begin{equation}
    \theta_i = \arccos(s'_i), \quad -1 \leq i \leq 1, \quad s_i' \in SI'
\end{equation}
\begin{equation}
    r=\frac{t_i}{N},t_i\in N
\end{equation}

Here, \( t \) represents the timestamp, and the interval \([0, 1]\) is divided into \( N \) equal parts to regularize the span of the polar coordinates. The encoding mapping in Equation (2) has two important properties. Firstly, it is bijective, as \(\cos(\Theta)\) is monotonically decreasing in \(\Theta \in [0, \pi]\), meaning that for a given time series, there is a unique corresponding value in polar coordinates, and its inverse mapping is also unique. Secondly, unlike Cartesian coordinates, polar coordinates maintain the absolute time relationship.

After mapping the one-dimensional signal to polar coordinates, we can easily utilize the angular perspective to identify temporal correlations within different time intervals by considering the trigonometric functions between each point. The Gramian Angular Field (GAF) is defined as:
\begin{equation}
G = \begin{bmatrix}
\cos(\theta_1 + \theta_1) & \ldots & \cos(\theta_1 + \theta_n) \\
\cos(\theta_2 + \theta_1) & \ldots & \cos(\theta_2 + \theta_n) \\
\vdots & \vdots & \vdots \\
\cos(\theta_n + \theta_1) & \ldots & \cos(\theta_n + \theta_n)
\end{bmatrix}\\
\end{equation}
\begin{equation}
    G= SI' \cdot SI - {\sqrt{I - SI^2}}' \cdot \sqrt{I - SI^2}
\end{equation}

where $I$ is a unit row vector $[1, 1, \ldots, 1]$, and $\theta_i$ ($i=1, \ldots, n$) is the angle between two vectors.After mapping the one-dimensional time series to polar coordinates, each time series of each step length is treated as a one-dimensional metric space. Since GAF is more sparse, the inner product is redefined in Cartesian coordinates as
\begin{equation}
   \langle \mathbf{x}, \mathbf{y} \rangle = x \cdot y - \sqrt{1 - x^2} \cdot \sqrt{1 - y^2} 
\end{equation}
relative to the traditional inner product, the new inner product includes a penalty term, making it easier to distinguish the required output from Gaussian noise. \( G \) is a Gram matrix defined as:
\begin{equation}
   G = \begin{bmatrix}
\langle s'_1, s'_1 \rangle & \ldots & \langle s'_1, s_n \rangle \\
\langle s'_2, s'_1 \rangle & \ldots & \langle s'_2, s_n \rangle \\
\vdots & \ddots & \vdots \\
\langle s_n, s_1 \rangle & \ldots & \langle s_n, s_n \rangle
\end{bmatrix}  
\end{equation}

The process of generating GAF is illustrated in the Fig\ref{fig:GAF}.
GAF incorporates temporal correlations, where the sum of $G_{(i,j \, | \, |,\ i-j=k)}$
over time intervals \( k \) is superimposed, and interpreted in terms of relative correlations. When \( k = 0 \), the main diagonal of $G_{i,i}$ consists of the raw values of the scaled time series. Utilizing the main diagonal, high-level features learned by deep neural networks can be used to approximate the reconstruction of the time series.

\subsection{Multi-level Feature Extraction}
As mentioned earlier, there is an imbalance in the long-range and short-range temporal relationships in EEG signals. Due to the basic architecture of GRC-Net being based on the ResNet architecture, its inherent spatial locality and translational invariance enhance the local modeling capability, thereby improving the ability to learn short-range temporal relationships. To model global information, we introduce the concept of hierarchical feature extraction to address the modeling of long-range temporal relationships.
\begin{figure*}[!t]
    \centering
    \setlength{\tabcolsep}{0pt} 
    \renewcommand{\arraystretch}{0} 
    \begin{minipage}{0.19\textwidth}
        \centering
        \includegraphics[height=0.9\linewidth]{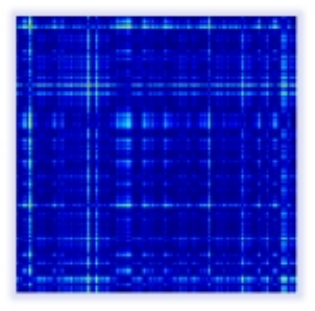}
    \end{minipage}
    \begin{minipage}{0.19\textwidth}
        \centering
        \includegraphics[height=0.9\linewidth]{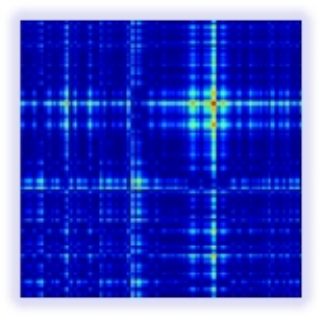}
    \end{minipage}
    \begin{minipage}{0.19\textwidth}
        \centering
        \includegraphics[height=0.9\linewidth]{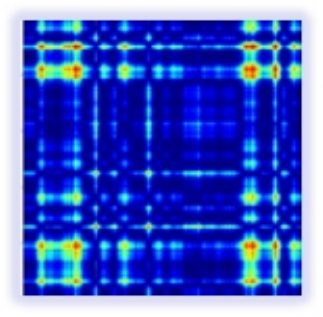}
    \end{minipage}
    \begin{minipage}{0.19\textwidth}
        \centering
        \includegraphics[height=0.9\linewidth]{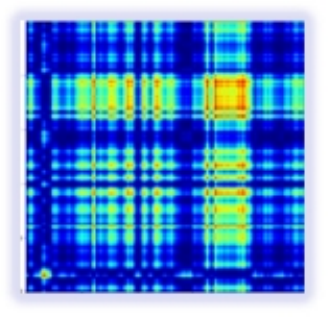}
    \end{minipage}
    \begin{minipage}{0.19\textwidth}
        \centering
        \includegraphics[height=0.9\linewidth]{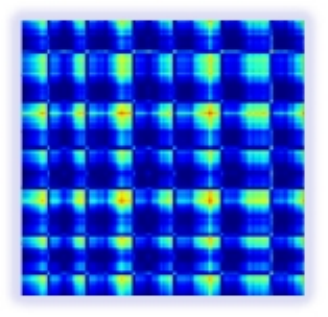}
    \end{minipage}
    \centering
    \begin{minipage}{0.19\textwidth}
        \centering
        \includegraphics[height=0.9\linewidth]{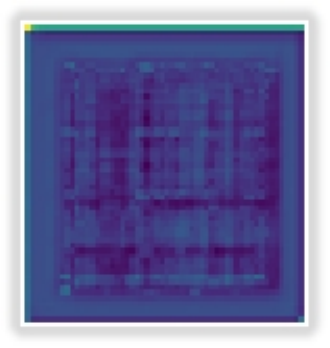}
    \end{minipage}
    \begin{minipage}{0.19\textwidth}
        \centering
        \includegraphics[height=0.9\linewidth]{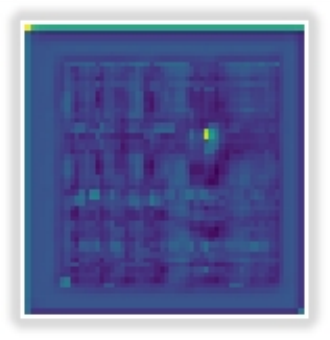}
    \end{minipage}
    \begin{minipage}{0.19\textwidth}
        \centering
        \includegraphics[height=0.9\linewidth]{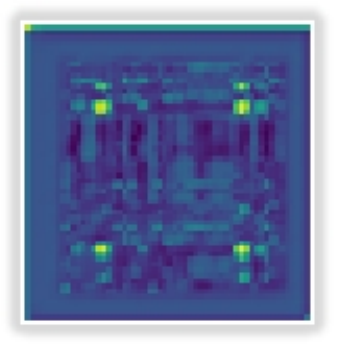}
    \end{minipage}
    \begin{minipage}{0.19\textwidth}
        \centering
        \includegraphics[height=0.9\linewidth]{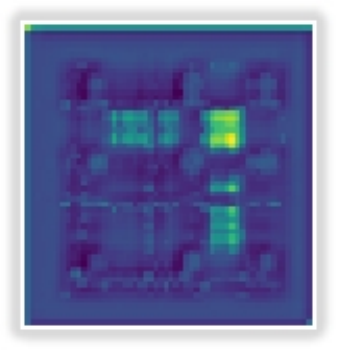}
    \end{minipage}
    \begin{minipage}{0.19\textwidth}
        \centering
        \includegraphics[height=0.9\linewidth]{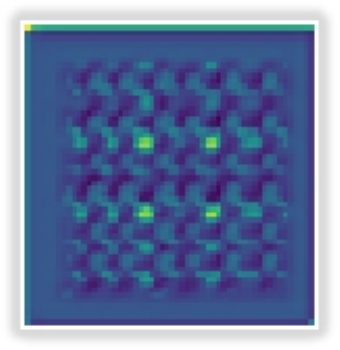}
    \end{minipage}
    \centering
    \begin{minipage}{0.19\textwidth}
        \centering
        \includegraphics[height=0.9\linewidth]{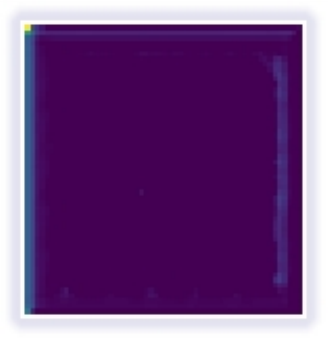}
    \end{minipage}
    \begin{minipage}{0.19\textwidth}
        \centering
        \includegraphics[height=0.9\linewidth]{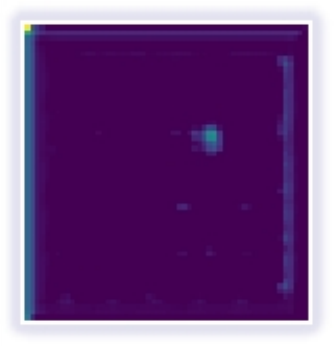}
    \end{minipage}
    \begin{minipage}{0.19\textwidth}
        \centering
        \includegraphics[height=0.9\linewidth]{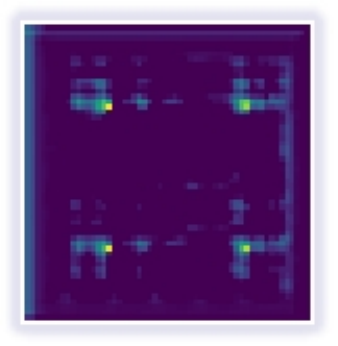}
    \end{minipage}
    \begin{minipage}{0.19\textwidth}
        \centering
        \includegraphics[height=0.9\linewidth]{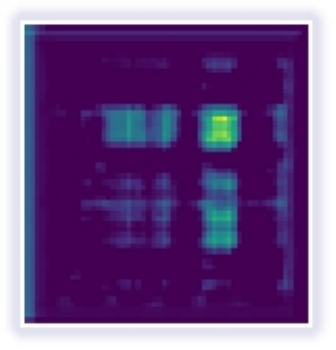}
    \end{minipage}
    \begin{minipage}{0.19\textwidth}
        \centering
        \includegraphics[height=0.9\linewidth]{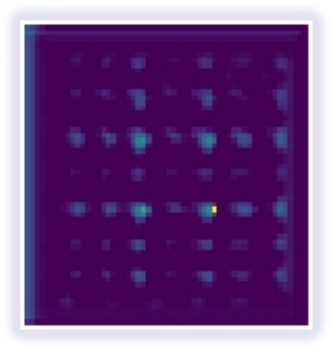}
    \end{minipage}
     \caption{The visualization of the original image, the feature map after primary feature extraction, and the feature map after cotattention can be observed, with each row representing five different classes from the Bonn dataset from left to right. It is evident that the integration of cotattention not only captures crucial information from a global perspective but also filters out some noise, effectively enhancing global feature extraction accuracy.}
\end{figure*}

The first step involves using Coattention to capture the long-range temporal relationships. For a given input \( X_{\text{in}} \in \mathbb{R}^{H \cdot W \cdot 3} \), we first use shallow feature extraction to increase the feature dimension, learn more feature information, and obtain feature maps \( X' (\mathbb{R}^{H \cdot W \cdot 3}) \). Subsequently, we pass \( X' \) into the CoT layer, where coattention is employed to further learn contextual key features, enabling global modeling.

Faced with the input feature maps, the CoT layer extracts static and dynamic context features in parallel. The CoT layer learns static context keys by encoding input keys using a 3x3 convolutional layer. 
\begin{equation}
    \text{Static} = \text{Conv}3 \times 3(\text{Input})
\end{equation}

The static context ensures the local features learned by taking advantage of the spatial locality and translational invariance of the convolution. In the dynamic context branch, the relationship between each key and query is used to fuse them and obtain the attention weight matrix between the key and query. 

\begin{equation}
    QK = \text{Cat}([\text{Static}, \text{Input}])
\end{equation}
\begin{equation}
    V = \text{Conv}1 \times 1(\text{Input})
\end{equation}

This matrix is then multiplied with the value to achieve a dynamic contextual representation of the input. This process fully integrates the information provided by the keys, queries, and values, enabling dynamic exploration of the global context.

\begin{equation}
    \text{Dynamic} = \text{LocalConvolution}[\text{Conv}1 \times 1(QK), V]
\end{equation}

Finally, the dynamic and static contexts are fused to facilitate self-attention learning. This design focuses more on learning rich contexts between adjacent keys than the traditional self-attention mechanism. Simultaneously, this enables the dynamic context to enhance the learning of visual representations under the guidance of the static context, effectively mitigating the limiting effect of the convolutional layer's receptive field on learning global features.

\begin{equation}
    \text{Output} = \text{Conv}1 \times 1(\text{Cat}[\text{Static}, \text{Dynamic}])
\end{equation}

\section{Experiment}
\subsubsection{BONN}
The EEG signals dataset originates from Bonn University \cite{andrzejak2001indications}, comprising data from five healthy individuals and five epileptic patients, constituting five subsets denoted as F, S, N, Z, and O. Each subset includes 100 data segments, each lasting for 23.6 seconds with 4097 data points. The signal resolution is 12 bits, with a sampling frequency of 173.61 Hz. The dataset details are summarized in the table \ref{tab:dataset}.

\begin{figure}
    \centering
    \includegraphics[width=1\columnwidth]{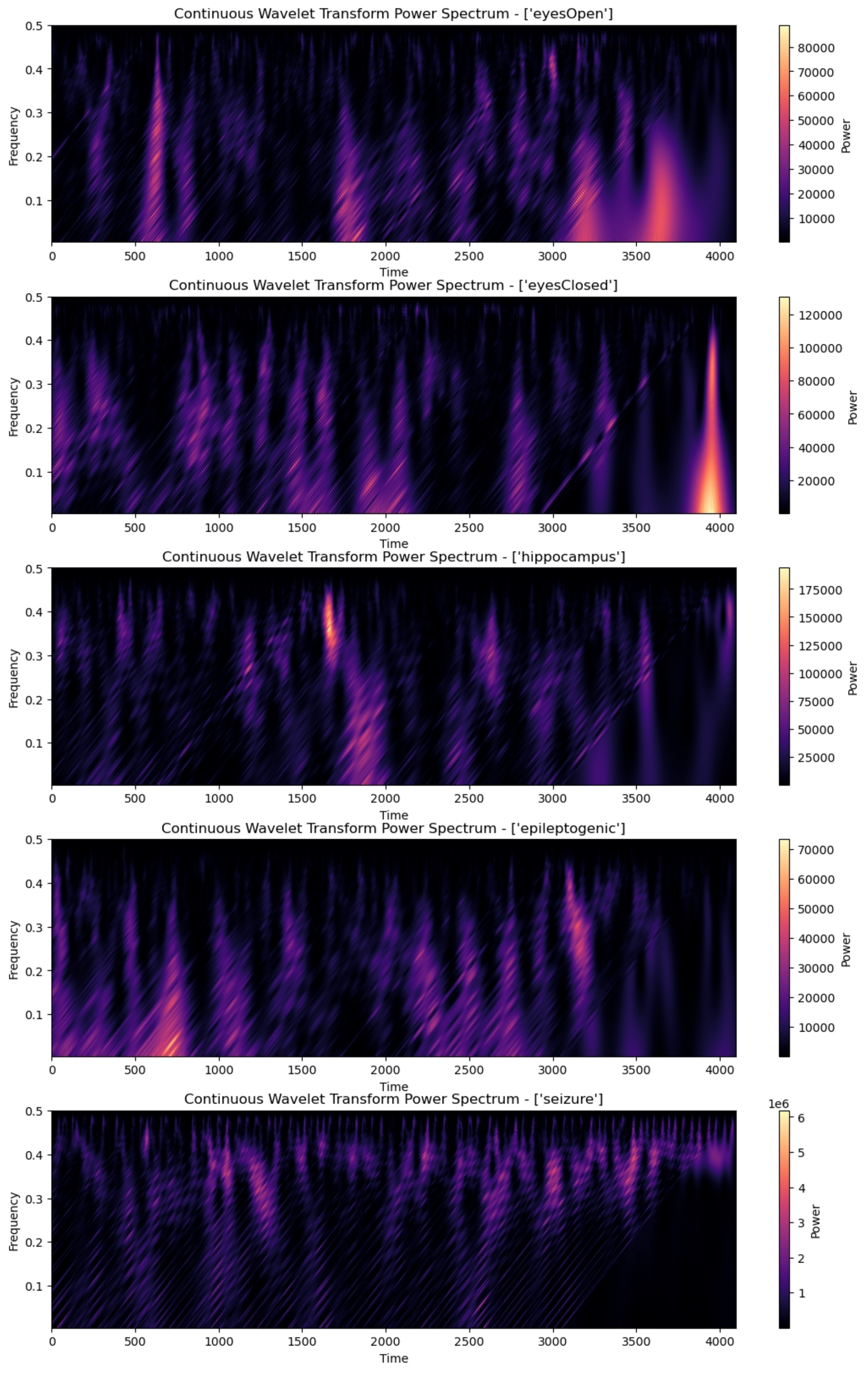}
    \caption{The five signals of the BoNN dataset are subjected to wavelet transform in this image, which simultaneously presents the frequency-domain and time-domain information of the signals while reducing noise. This visualization, compared to simple one-dimensional signals, is more conducive to analysis.}
    \label{fig:BONN visual}
\end{figure}

\begin{table}
\setlength\tabcolsep{2pt} 
  \centering
  \caption{The description of Bonn Dataset}
  \begin{tabularx}{\columnwidth}{p{1.2cm} *{5}{>{\centering\arraybackslash}X} } 
    \toprule
    Model & \multicolumn{2}{c}{Healthy} & \multicolumn{3}{c}{Epilepsy Patients} \\
    \cmidrule(lr){2-3} \cmidrule(lr){4-6}
    & Z & O & N & F & S \\
    \midrule
    Status &Eyes Open&Eyes Closed&Inter--ictal&Inter--ictal&Ictal\\
    Data Type&Scalp&Scalp&Intra--cranial&Intra--cranial&Intra--cranial\\
    Electrode Placement&Scalp&Scalp&Hippo--campal Region&Lesion Area&Lesion Area\\
    \bottomrule
  \end{tabularx}
  \label{tab:dataset}
\end{table}

\subsection{Comparison with state-of-the-art methods}
Previous studies have explored various combinations of predictions on the BoNN dataset, such as binary, ternary, and quaternary classification tasks, each with different internal configurations. However, there has been limited targeted prediction specifically for the five-class classification task due to its inherent difficulty. Therefore, we compared the performance of our proposed GRC-Net on the BoNN dataset with other models. Table 1 illustrates the performance differences among different models for the five-class classification task. It is observed that traditional classification networks like AlexNet and VGG achieve an accuracy of 81.15\%. Despite the recent advancements in network models, the Improved CNN model, which utilizes a preprocessing mechanism involving adaptive rate sampling, modified activity selection, filtering, and wavelet decomposition to extract highly discriminatory features, achieves an accuracy of 92\%, indicating room for further enhancement. The Ensemble CNN model, which utilizes overlapping segments of EEG signals, achieves an accuracy of 93\%. Two models proposed in 2023 (IVMD-WmRMR-DF,DARLNet) show accuracies of 89.33\% and 90.17\%, respectively, without significant advantages, but their signal processing methods still hold innovative value. In comparison, our model synthesizes previous work, drawing inspiration and innovating with new approaches, resulting in an accuracy of 93.66\%, recall of 93.24\%, precision of 93.16\%, and F1-score of 93.14\%. The comparison highlights the significant advantages of our approach.
\begin{table}
\setlength\tabcolsep{1.8pt}
  \centering
  \caption{The comparison between our proposed model and other models}
  \begin{tabularx}{\columnwidth}{X *{5}{c}}
    \toprule
    Model & Rec(\%)↑ & Acc(\%)↑ & Prec(\%)↑ & F1(\%)↑ & Dataset \\
    \midrule
    AlexNet\cite{QIU2023104652} &79.24&79.48&79.37&79.12&\\
    Vgg16\cite{QIU2023104652} &81.15&80.52&80.41&80.03&\\
    1DCNN-LSTM\cite{QIU2023104652} &81.70 &82.00&81.78&81.56&\\
    1DCNN\cite{acharya2018deep}&-&88.70&-&-&\\
    1D-TP+BT\cite{kaya2018stable}&-&82.20&-&-&\\
    Double-DNN\cite{Chen2018CostSensitiveDA}&81.70&82.00& 81.70&81.56&\\
    Improved CNN\cite{hussain2022epileptic}&92.00&-&-&92.70&\\
    Ensemble CNN\cite{kim2021epileptic}&-&93.00&-&-&\\
    IVMD-WmRMR-DF\cite{QIN2023104644} & 89.33 & 89.33 & 89.6 & 89.13 & \\
    DARLNet\cite{QIU2023104652} & 90.16 & 90.17 & 90.00 & 90.05 & \\
    \textbf{Proposed} & 93.24 & 93.66 & 93.16 & 93.14\\
    \midrule
  \end{tabularx}
  \label{tab:COM}
\end{table}
\subsubsection{Ablation Experiment}
Table \ref{tab:COM} conducted ablation experiments on the components of GRC-Net. Compared to the complete model, the performance of the model improved by approximately 7\% in the one-dimensional signal transformation process using the RM method. This once again demonstrates that traditional methods of transforming one-dimensional signals into two-dimensional signals result in the loss of temporal dependency information, thereby affecting model performance. Regarding hierarchical extraction, removing the coattention mechanism for extracting global features leads to a performance drop of around 2\%, confirming the importance of the model in extracting features from different scales. Finally, we removed the inception architecture, noting that the model's input remained two-dimensional images without RM processing. We replaced the original inception architecture with the VGG architecture, aiming to maintain the concept of hierarchical feature extraction. It is noteworthy that the RU architecture excels in extracting local features more effectively.
\begin{table}
\setlength\tabcolsep{2pt}
  \centering
  \caption{The comparison between our proposed model and other models}
  \begin{tabularx}{\columnwidth}{X *{5}{c}}
    \toprule
    Model & Rec(\%)↑ & Acc(\%)↑ & Prec(\%)↑ & F1(\%)↑ & Dataset \\
    \midrule
    Without GM & 83.12 & 83.62 & 83.34 & 83.22 & \\
    Without cotattention & 90.80 & 91.60 & 90.45 & 96.28 & \\
    without RU &79.83&80.16&79.42&79.62\\
    \midrule
  \end{tabularx}
  \label{tab:COM}
\end{table}

\subsection{Limitations}
Although our research has achieved excellent performance, there are still research gaps that need to be addressed in future studies. Firstly, we focused on the five-class classification problem, which, while more complex compared to binary and ternary classification tasks, requires further experimental validation. Secondly, our experiments primarily concentrated on the BoNN dataset, without performance validation on other datasets. Lastly, the Gram Matrix (GM) algorithm can be divided into two transformation methods: summation and difference. This study only utilized the summation method and did not explore the effects of signal transformation using the difference method.
\section{Conclusion}
In conclusion, addressing the classification challenges in epilepsy diagnosis and treatment, we have proposed a novel model architecture, GCN-Net, which leverages the Gram matrix to transform one-dimensional signals into two-dimensional signals while preserving the temporal dependencies during the transformation process. Furthermore, we have introduced a hierarchical feature extraction concept, utilizing coattention to capture global key features and employing inception for extracting key features of local information, resulting in an improved five-class classification accuracy on the BoNN dataset.

Through our research and methodologies, we can enhance the diagnostic accuracy and treatment effectiveness of epilepsy. Our work provides valuable insights and directions for future epilepsy research and clinical practices, potentially offering improved medical services and care for patients. With ongoing technological advancements and deeper research, we believe that we can continuously refine diagnostic and treatment approaches for epilepsy, ultimately delivering a brighter future for patients.

\bibliographystyle{spmpsci}
\bibliography{Article} 

\begin{thebibliography}{10}
\providecommand{\url}[1]{{#1}}
\providecommand{\urlprefix}{URL }
\expandafter\ifx\csname urlstyle\endcsname\relax
  \providecommand{\doi}[1]{DOI~\discretionary{}{}{}#1}\else
  \providecommand{\doi}{DOI~\discretionary{}{}{}\begingroup \urlstyle{rm}\Url}\fi

\bibitem{acharya2018deep}
Acharya, U.R., Oh, S.L., Hagiwara, Y., Tan, J.H., Adeli, H.: Deep convolutional neural network for the automated detection and diagnosis of seizure using eeg signals.
\newblock Computers in biology and medicine \textbf{100}, 270--278 (2018)

\bibitem{andrzejak2001indications}
Andrzejak, R.G., Lehnertz, K., Mormann, F., Rieke, C., David, P., Elger, C.E.: Indications of nonlinear deterministic and finite-dimensional structures in time series of brain electrical activity: Dependence on recording region and brain state.
\newblock Physical Review E \textbf{64}(6), 061,907 (2001)

\bibitem{cai2018reconstruction}
Cai, L., Wei, X., Wang, J., Yu, H., Deng, B., Wang, R.: Reconstruction of functional brain network in alzheimer's disease via cross-frequency phase synchronization.
\newblock Neurocomputing \textbf{314}, 490--500 (2018)

\bibitem{Chen2018CostSensitiveDA}
Chen, X., Ji, J., Ji, T., Li, P.: Cost-sensitive deep active learning for epileptic seizure detection.
\newblock Proceedings of the 2018 ACM International Conference on Bioinformatics, Computational Biology, and Health Informatics  (2018).
\newblock \urlprefix\url{https://api.semanticscholar.org/CorpusID:52095075}

\bibitem{chen2012towards}
Chen, Y.N., Chang, K.m.K., Mostow, J.: Towards using eeg to improve asr accuracy.
\newblock In: Proceedings of the 2012 Conference of the North American Chapter of the Association for Computational Linguistics: Human Language Technologies, pp. 382--385 (2012)

\bibitem{Chou2023ConvolutionalNN}
Chou, C.H., Shen, T.W., Tung, H., Hsieh, P.F., Kuo, C.E., Chen, T.M., Yang, C.W.: Convolutional neural network-based fast seizure detection from video electroencephalograms.
\newblock Biomed. Signal Process. Control. \textbf{80}, 104,380 (2023).
\newblock \urlprefix\url{https://api.semanticscholar.org/CorpusID:253753356}

\bibitem{Du2024ElectroencephalographicSD}
Du, X., Wang, X., Zhu, L., Ding, X., Lv, Y., Qiu, S., Liu, Q.: Electroencephalographic signal data augmentation based on improved generative adversarial network.
\newblock Brain Sciences  (2024).
\newblock \urlprefix\url{https://api.semanticscholar.org/CorpusID:269063915}

\bibitem{frantzidis2014functional}
Frantzidis, C.A., Vivas, A.B., Tsolaki, A., Klados, M.A., Tsolaki, M., Bamidis, P.D.: Functional disorganization of small-world brain networks in mild alzheimer's disease and amnestic mild cognitive impairment: an eeg study using relative wavelet entropy (rwe).
\newblock Frontiers in aging neuroscience \textbf{6}, 224 (2014)

\bibitem{Gao2021GenerativeAN}
Gao, B., Zhou, J., Yang, Y., Chi, J., Yuan, Q.: Generative adversarial network and convolutional neural network-based eeg imbalanced classification model for seizure detection.
\newblock Biocybernetics and Biomedical Engineering  (2021).
\newblock \urlprefix\url{https://api.semanticscholar.org/CorpusID:244745679}

\bibitem{hussain2022epileptic}
Hussain, S.F., Qaisar, S.M.: Epileptic seizure classification using level-crossing eeg sampling and ensemble of sub-problems classifier.
\newblock Expert Systems with Applications \textbf{191}, 116,356 (2022)

\bibitem{jia2022efficient}
Jia, M., Liu, W., Duan, J., Chen, L., Chen, C.P., Wang, Q., Zhou, Z.: Efficient graph convolutional networks for seizure prediction using scalp eeg.
\newblock Frontiers in Neuroscience \textbf{16}, 967,116 (2022)

\bibitem{Kantipudi2024AnIG}
Kantipudi, M.P., Kumar, N.S.P., Aluvalu, D.R., Selvarajan, S., Kotecha, K.: An improved gbso-taenn-based eeg signal classification model for epileptic seizure detection.
\newblock Scientific Reports \textbf{14} (2024).
\newblock \urlprefix\url{https://api.semanticscholar.org/CorpusID:266845628}

\bibitem{kaya2018stable}
Kaya, Y., Ertu{\u{g}}rul, {\"O}.F.: A stable feature extraction method in classification epileptic eeg signals.
\newblock Australasian physical \& engineering sciences in medicine \textbf{41}, 721--730 (2018)

\bibitem{kim2021epileptic}
Kim, M.K.: Epileptic seizure detection using cnn ensemble models based on overlapping segments of eeg signals.
\newblock KIPS Transactions on Software and Data Engineering \textbf{10}(12), 587--594 (2021)

\bibitem{Lee2024ARH}
Lee, D., Kim, B., Kim, T., Joe, I., Chong, J., Min, K., Jung, K.: A resnet-lstm hybrid model for predicting epileptic seizures using a pretrained model with supervised contrastive learning.
\newblock Scientific Reports \textbf{14} (2024).
\newblock \urlprefix\url{https://api.semanticscholar.org/CorpusID:266997133}

\bibitem{Lu2023AnES}
Lu, X., Wen, A., Sun, L., Wang, H., Guo, Y., Ren, Y.: An epileptic seizure prediction method based on cbam-3d cnn-lstm model.
\newblock IEEE Journal of Translational Engineering in Health and Medicine \textbf{11}, 417 -- 423 (2023).
\newblock \urlprefix\url{https://api.semanticscholar.org/CorpusID:259368814}

\bibitem{Mishra2022ADB}
Mishra, S., Satapathy, S.K., Mohanty, S.N., Pattnaik, C.R.: A dm-elm based classifier for eeg brain signal classification for epileptic seizure detection.
\newblock Communicative \& Integrative Biology \textbf{16} (2022).
\newblock \urlprefix\url{https://api.semanticscholar.org/CorpusID:254762951}

\bibitem{oh2019deep}
Oh, S.L., Vicnesh, J., Ciaccio, E.J., Yuvaraj, R., Acharya, U.R.: Deep convolutional neural network model for automated diagnosis of schizophrenia using eeg signals.
\newblock Applied Sciences \textbf{9}(14), 2870 (2019)

\bibitem{Panda2023HybridWO}
Panda, S., Mishra, S., Mohanty, M.N.: Hybrid wca–pso optimized ensemble extreme learning machine and wavelet transform for detection and classification of epileptic seizure from eeg signals.
\newblock Augmented Human Research \textbf{8}, 1--12 (2023).
\newblock \urlprefix\url{https://api.semanticscholar.org/CorpusID:266150699}

\bibitem{petrantonakis2009emotion}
Petrantonakis, P.C., Hadjileontiadis, L.J.: Emotion recognition from eeg using higher order crossings.
\newblock IEEE Transactions on information Technology in Biomedicine \textbf{14}(2), 186--197 (2009)

\bibitem{Qi2023PredictingES}
Qi, N., Piao, Y., Yu, P., Tan, B.: Predicting epileptic seizures based on eeg signals using spatial depth features of a 3d-2d hybrid cnn.
\newblock Medical \& Biological Engineering \& Computing \textbf{61}, 1845--1856 (2023).
\newblock \urlprefix\url{https://api.semanticscholar.org/CorpusID:257694139}

\bibitem{QIN2023104644}
Qin, X., Xu, D., Dong, X., Cui, X., Zhang, S.: Eeg signal classification based on improved variational mode decomposition and deep forest.
\newblock Biomedical Signal Processing and Control \textbf{83}, 104,644 (2023).
\newblock \doi{https://doi.org/10.1016/j.bspc.2023.104644}.
\newblock \urlprefix\url{https://www.sciencedirect.com/science/article/pii/S1746809423000770}

\bibitem{QIU2023104652}
Qiu, X., Yan, F., Liu, H.: A difference attention resnet-lstm network for epileptic seizure detection using eeg signal.
\newblock Biomedical Signal Processing and Control \textbf{83}, 104,652 (2023).
\newblock \doi{https://doi.org/10.1016/j.bspc.2023.104652}.
\newblock \urlprefix\url{https://www.sciencedirect.com/science/article/pii/S174680942300085X}

\bibitem{Singh2023AutomaticPO}
Singh, Y.P., Lobiyal, D.K.: Automatic prediction of epileptic seizure using hybrid deep resnet-lstm model.
\newblock AI Commun. \textbf{36}, 57--72 (2023).
\newblock \urlprefix\url{https://api.semanticscholar.org/CorpusID:256265308}

\bibitem{Sivasaravanababu2021AnEE}
Sivasaravanababu, S., Prabhu, V.G., Parthasarathy, V., Mahendran, R.K.: An efficient epileptic seizure detection based on tunable q-wavelet transform and dcvae-stacked bi-lstm model using electroencephalogram.
\newblock The European Physical Journal Special Topics \textbf{231}, 2425 -- 2437 (2021).
\newblock \urlprefix\url{https://api.semanticscholar.org/CorpusID:245478026}

\bibitem{Swetha2023OppositionalCS}
Swetha, K.R., Gayatri, P., Kumar, E., N., S., Sundaram, S.M.: Oppositional crow search algorithm based artificial neural network for epileptic seizure classification.
\newblock 2023 International Conference on Evolutionary Algorithms and Soft Computing Techniques (EASCT) pp. 1--5 (2023).
\newblock \urlprefix\url{https://api.semanticscholar.org/CorpusID:267124154}

\bibitem{Thuwajit2022EEGWaveNetMC}
Thuwajit, P., Rangpong, P., Sawangjai, P., Autthasan, P., Chaisaen, R., Banluesombatkul, N., Boonchit, P., Tatsaringkansakul, N., Sudhawiyangkul, T., Wilaiprasitporn, T.: Eegwavenet: Multiscale cnn-based spatiotemporal feature extraction for eeg seizure detection.
\newblock IEEE Transactions on Industrial Informatics \textbf{18}, 5547--5557 (2022).
\newblock \urlprefix\url{https://api.semanticscholar.org/CorpusID:245095799}

\bibitem{Tuncer2021EpilepsyAR}
Tuncer, T., Dogan, S., Naik, G.R., Plawiak, P.: Epilepsy attacks recognition based on 1d octal pattern, wavelet transform and eeg signals.
\newblock Multimedia Tools and Applications \textbf{80}, 25,197 -- 25,218 (2021).
\newblock \urlprefix\url{https://api.semanticscholar.org/CorpusID:234802315}

\bibitem{wusthoff2009limitations}
Wusthoff, C., Shellhaas, R., Clancy, R.: Limitations of single-channel eeg on the forehead for neonatal seizure detection.
\newblock Journal of Perinatology \textbf{29}(3), 237--242 (2009)

\bibitem{ZHOU2024108086}
Zhou, Q., Zhang, S., Du, Q., Ke, L.: Rihanet: A residual-based inception with hybrid-attention network for seizure detection using eeg signals.
\newblock Computers in Biology and Medicine \textbf{171}, 108,086 (2024).
\newblock \doi{https://doi.org/10.1016/j.compbiomed.2024.108086}.
\newblock \urlprefix\url{https://www.sciencedirect.com/science/article/pii/S0010482524001707}

\end{thebibliography}
\end{document}